\pdfoutput=1

\documentclass[11pt]{article}

\usepackage[preprint]{acl}

\usepackage{times}
\usepackage{latexsym}

\usepackage[T1]{fontenc}

\usepackage[utf8]{inputenc}

\usepackage{microtype}

\usepackage{inconsolata}

\usepackage{graphicx}
\usepackage{float}

\usepackage{amsmath}
\usepackage{amssymb}
\usepackage{amsthm}
\usepackage{booktabs}
\usepackage{longtable}
\usepackage{placeins} 
\usepackage{makecell}
\usepackage{multirow}
\usepackage{makecell}

\usepackage[capitalize]{cleveref}
\usepackage{array}
\newcolumntype{P}[1]{>{\centering\arraybackslash}p{#1}}

\usepackage{cleveref}

\usepackage{fontawesome}

\usepackage{adjustbox}

\usepackage{tabularx}

\theoremstyle{definition}

\crefname{definition}{Definition}{Definitions}%
\crefname{section}{Sec.}{Secs.}%

\usepackage{scrextend} 
\usepackage{siunitx}

\newcommand{\err}[1]{\tiny{$\pm$}\scriptsize{\num[round-mode=places,round-precision=1]{#1}}}
\usepackage{makecell}
\newcommand{\capitalizefirst}[1]{\MakeUppercase{\expandafter\@car#1\@nil}\expandafter\@cdr#1\@nil}

\usepackage{changepage}   

\newcommand{\methodlong}{explanation-consistency finetuning}
\newcommand{\method}{EC-finetuning}
\newcommand{\methodshort}{EC}

\title{Towards Consistent Natural-Language Explanations\\via Explanation-Consistency Finetuning}

\author{Yanda Chen$^{1,2}$, Chandan Singh$^2$, Xiaodong Liu$^2$, Simiao Zuo$^2$,\\\textbf{Bin Yu$^3$, He He$^4$, Jianfeng Gao$^2$}
\\
\\$^1$Columbia University, $^2$Microsoft Research, $^3$University of California, Berkeley, $^4$NYU
}

\begin{document}

\maketitle

\begin{abstract}
Large language models (LLMs) often generate convincing, fluent explanations.
However, different from humans, 
they often generate \textit{inconsistent} explanations
on different inputs.
For example, an LLM may generate the explanation ``\textit{all birds can fly}'' when answering the question ``\textit{Can sparrows fly?}'' but meanwhile answer ``\textit{no}'' to the related question ``\textit{Can penguins fly?}''.
Explanations should be consistent across related examples 
so that
they allow a human to simulate the LLM's decision process on multiple examples.

We propose \textbf{\methodlong} (\method), a method that adapts LLMs to generate more consistent natural-language explanations
on related examples.
\method{} involves finetuning LLMs on 
synthetic
data that is carefully constructed 
to contain consistent explanations.
Across a variety of question-answering datasets in various domains, \method{} yields a \textbf{10.0\%} relative explanation consistency improvement
on four finetuning datasets,
and generalizes to seven out-of-distribution datasets not seen during finetuning (\textbf{+4.5\%} relative). 
Code is available at \url{https://github.com/yandachen/explanation-consistency-finetuning}.
\end{abstract}

\section{Introduction}

Pre-trained large language models (LLMs) have shown impressive proficiency in a range of complex natural language processing tasks, significantly advancing the field and opening new frontiers for applications~\cite{brown2020language,touvron2023llama2,openai2023gpt4}.
However, the opaqueness of these models' decision making process has hindered their use in high-stakes applications such as healthcare, and raised issues related to 
regulatory pressure, safety, and alignment~\cite{goodman2016european,amodei2016concrete,gabriel2020artificial}.
Moreover, this lack of interpretability has heavily limited the use of LLMs in fields such as social science and data analysis~\cite{ziems2023can}, where trustworthy interpretation (rather than model deployment) is itself the end goal.

\begin{figure}[t!]
    \includegraphics[width=\columnwidth]{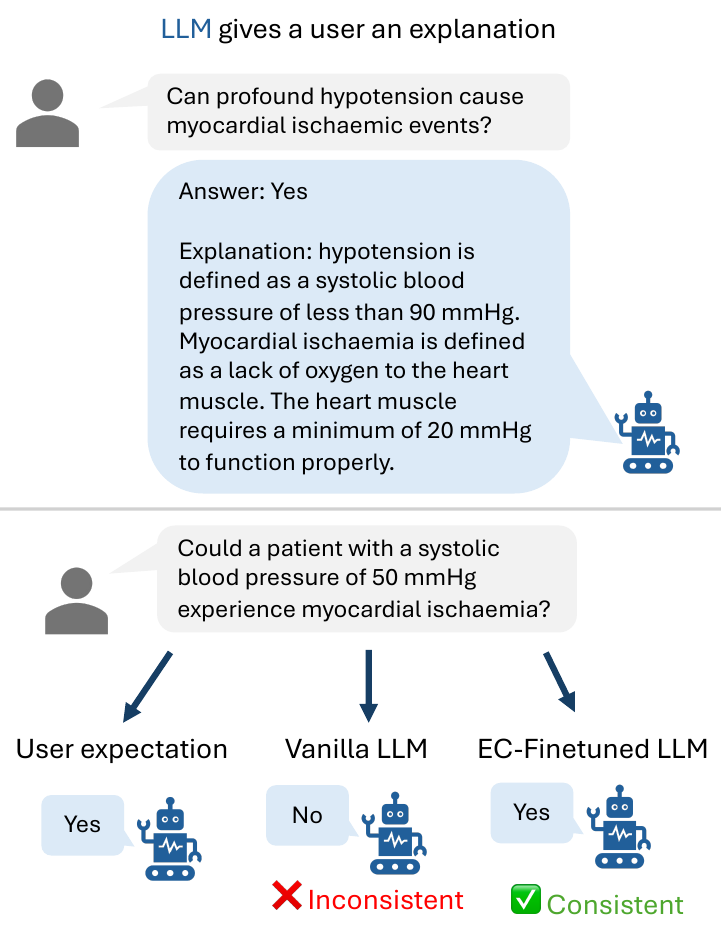}
    \caption{\method{} adapts an LLM to provide explanations that are more consistent with a user's expectation of LLM answers across related questions.}
    \label{fig:intro}
\end{figure}

\begin{figure*}[t!]
    \includegraphics[width=\textwidth]{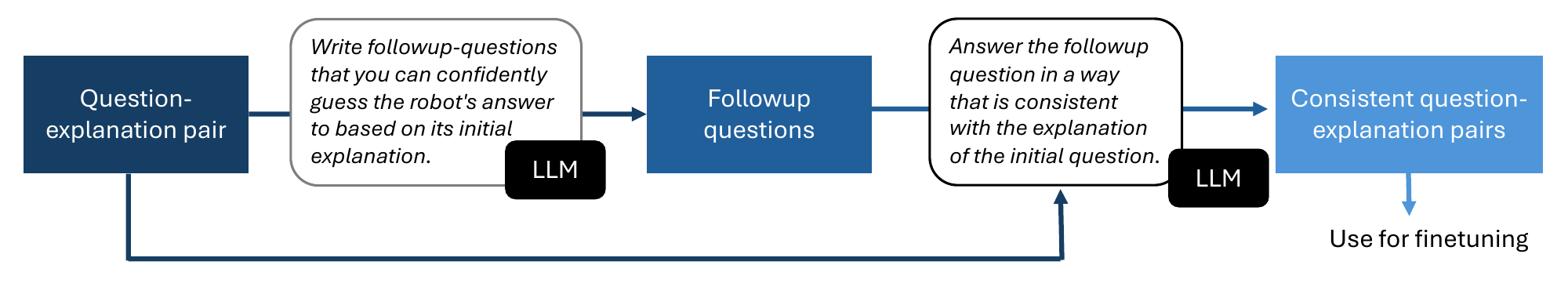}
    \caption{\method{} synthetically augments the examples in a dataset using LLMs.
    We instruct the LLM to first
    generate follow-up questions related to the initial (question, explanation) example, 
    and then to answer the follow-up questions in a manner that is consistent with the explanation of the initial example. 
    }
    \label{fig:method}
\end{figure*}

One growing avenue into interpretability is natural-language explanations produced by LLMs.
These explanations are potentially very powerful,
helping users predict model behavior \cite{johnson1980mental,
bansal2019beyond}, which is useful to
calibrate a model's capacity and limitations,
e.g. limiting its demographic bias~\cite{NEURIPS2020_92650b2e}.
However, these natural-language explanations are limited by the inherent 
inconsistency
of LLMs.
For example, one recent work 
finds that modern LLMs often generate inconsistent explanations and answers on
very related questions~\cite{chen2023models}.
In fact, LLMs often even struggle to consistently answer rephrasings of the same question~\cite{sclar2023quantifying,zhang2023tell}.
It is unclear if popular methods for adapting LLMs, e.g. supervised finetuning or reinforcement learning from human feedback, are able to solve this issue.

We address this issue by introducing \methodlong{} (\method).
\method{}  finetunes an LLM on synthetic data that is precisely constructed to contain consistent explanations.
We start with a question-explanation pair (e.g., ``\textit{Can sparrows fly?}'', ``\textit{all birds can fly}''), generate a set of related questions (e.g., ``\textit{Can penguins fly?}''), and then answer the related questions to be consistent with the initial explanation (e.g., ``{\textit{all birds can fly so penguins can fly}}'').
We generate the synthetic data by prompting LLMs, which can be the same as or different from the explanation LLM.

We apply \method{} to question-answering datasets and find that it improves 
the consistency of natural-language explanations of LLaMA2-13B
by \textbf{10.0\%} relative on four finetuning datasets,
and also generalizes to seven out-of-distribution datasets unseen during finetuning (\textbf{+4.5\%} relative). 
This suggests that \method{} may be generally useful for helping users build mental models of an LLM from its explanations (see \cref{fig:intro}).

\section{Related work}
\label{sec:related-work}

\paragraph{Generating and improving natural-language explanations}
A great deal of recent work has studied improving LLM controllability, e.g. through supervised finetuning~\cite{liu2019multi}, reinforcement learning from human feedback~\cite{ouyang2022training}
or learning from human explanations~\cite{stiennon2020learning}.
Two related works study the consistency in the generations made by an LLM, either between the generation and validation of LLMs~\cite{li2023benchmarking} or LLM predictions on implications of an original question~\cite{akyurek2024deductive}.
In contrast to \method, these works do not focus directly on improving an LLM's explanation capabilities.

Many works extend and analyze explanations given by chain-of-thought prompting~\cite{wei2022chain}, e.g. by evaluating counterfactuals introduced into the chain of thought~\cite{gat2023faithful}, testing their robustness to mistakes introduced into the reasoning chain~\cite{lanham2023measuring},
or using contrastive chain-of-thought to induce reliance on the reasoning chain~\cite{chia2023contrastive}.
These methods do not alter the underlying LLM, and thus can be used in conjunction with \method.

\paragraph{Evaluating natural-language explanations}
We summarize three existing orthogonal metrics for explanations: consistency, plausibility, and faithfulness. 
\textit{Consistency}, which we focus on in this work, measures if the model generates consistent explanations on similar examples \cite{hase-bansal-2020-evaluating, chen2023models}.
\textit{Plausibility} evaluates humans' preference of an explanation based on its factual correctness and logical coherence \cite{herman2017promise, lage2019evaluation, jacovi-goldberg-2020-towards}.
It is different from \textit{faithfulness}, which measures whether an explanation is consistent with the model's internal decision process \cite{harrington1985harvey,
jacovi-goldberg-2020-towards}.

\section{Method: \method}
\label{sec:method}
\paragraph{Explanation-consistency Finetuning}

\method{} is an intuitive method that 
augments data in a manner that enhances explanation consistency (\cref{fig:method}).
Specifically, it prompts LLMs to augment data in two steps.
In the first step, a question-explanation pair is given to an LLM (e.g., ``\textit{Can sparrows fly?}'', ``\textit{all birds can fly}''), with the task of generating follow-up questions related to the explanation of the initial question (e.g., ``\textit{Can penguins fly?}'').
This is achieved by explicitly prompting the LLM to generate questions that are answerable given the initial explanation.

In the second step, another LLM generates answers and explanations for the follow-up questions.
To ensure these answers and explanations are consistent with the explanation in the initial question, 
the initial question-explanation pair is presented in the prompt, alongside explicit instructions to keep the new explanation consistent with the initial (e.g., ``{\textit{all birds can fly so penguins can fly}}''.)
Finally, these augmented questions, along with their explanations and answers are used for finetuning an LLM to generate more consistent explanations.

We use different LLMs 
for the two data augmentation steps (here, \mbox{GPT-4}~\cite{openai2023gpt4} for the first step and Claude-2\footnote{https://www.anthropic.com/index/claude-2} for the second step) to avoid issues with LLMs that favor their own outputs \cite{zheng2023judging}. Precise prompts are given in \cref{sec:prompting_details}.
Note that the two data augmentation LLMs can be, but are not required to be, identical to the explanation LLM.

\paragraph{Measuring consistency}
Evaluating the consistency of model explanations is challenging.
Here, we follow the metric proposed by \citeauthor{chen2023models}~\citeyear{chen2023models}, which measures explanation consistency\footnote{What we call ``consistency'', \citeauthor{chen2023models} call ``counterfactual simulatability precision''.} as the fraction of answers on follow-up questions that match a human user's expectation based on an explanation (similar to \cref{fig:intro}); the metric ranges from 0 to 1, with 1 being a perfect score.
Additionally, following \citeauthor{chen2023models}~\citeyear{chen2023models} we use LLMs as a user simulator, as it was found to reliably emulate a user's predictions on follow-up questions.

We evaluate consistency on two types of follow-up questions: related questions and rephrased questions.
For related questions, similar to how we generate EC data, we again use GPT-4 to generate.
For rephrased questions, we prompt GPT-4 to generate exact paraphrases of the original questions.

To ensure this metric from \citeauthor{chen2023models}~\citeyear{chen2023models} is reliable, we conduct two sanity checks. First, we check if the metric is stable with respect to how the metric is computed. We run various perturbations (e.g. varying how the followup questions are generated or the explanation format) and find the metric to be quite stable~(see \cref{tab:consistency_stability}).
Second, we study the correlation between explanation consistency and explanation length to see if the metric can be easily hacked by generating shorter/longer explanations. We do not observe a correlation on any of the 7 unseen datasets~(see \cref{tab:consistency_length_correlation}). 

\section{Results}

\subsection{Experimental setup}

\begin{table}
    \centering
    \fontsize{9.5}{10.5}\selectfont
    \setlength{\tabcolsep}{2.7pt}
\begin{tabular}{llll}
\toprule
& Dataset & Domain & Skills \\
\midrule
\multirow{3}{*}{\rotatebox[origin=c]{90}{FT}} & StrategyQA & Open & Facts, Reason \\
&MedQA & Medical & Facts, Reason \\
&MedMCQA & Medical & Facts \\
\midrule
\multirow{7}{*}{\rotatebox[origin=c]{90}{Unseen}} &BoolQ & Open & Facts \\
&NQ & Open & Facts \\
&MS-Marco & Open & Facts \\
&OBQA & Open & Reason, \em{\textbf{Commonsense}} \\
&MMLU-Med & Medical & Facts \\
&PubMedQA & Medical & Facts, Reason, \em{\textbf{Quantitative}} \\
&ARC-Easy & \em{\textbf{Science}} & Facts, Reason \\
\bottomrule
\end{tabular}

    \caption{We show the domain and required skills of each dataset. New domains and skills in testing datasets unseen during training are bold and italicized.
    }
    \label{tab:datasets}
\end{table}

\begin{table*}[ht]
    \fontsize{10}{11.5}\selectfont
    \centering
    \begin{tabular}{ll|ll|ll|ll}
\toprule
& & \multicolumn{4}{c|}{Consistency} & \multicolumn{2}{c}{\multirow{2}{*}{Accuracy}} \\
\cmidrule[0.3pt]{3-6}
& & \multicolumn{2}{c}{Related Questions} & \multicolumn{2}{c|}{Rephrased Questions} & \multicolumn{2}{c}{} \\
\cmidrule[0.3pt]{3-8}
 &  &  Baseline &  \textbf{\methodshort}&  Baseline & \textbf{\methodshort} &  Baseline & \textbf{\methodshort}\\
 \midrule
 \parbox[c]{2mm}{\multirow{5}{*}{\rotatebox[origin=c]{90}{Finetuning}}} & 
   StrategyQA&  76.0\err{0.7} & 83.8\err{0.6} & -- & -- & 73.1\err{1.7} & 74.4\err{1.6} \\
 &  MedQA-Diff&  60.8\err{1.1} & 73.1\err{1.0} & -- & -- & 52.9\err{2.3} & 56.6\err{2.3} \\
 &  MedQA-Sim&  62.0\err{1.2} & 65.0\err{1.2} & -- & -- & 53.1\err{2.2} & 55.1\err{2.2} \\
 &  MedMCQA&  68.2\err{1.0} & 71.3\err{0.9} & -- & -- & 45.5\err{2.1} & 49.2\err{2.1} \\
& \textbf{AVG} & \textbf{66.7} & \textbf{73.3} & -- & -- & \textbf{56.1} & \textbf{58.8}\\
\midrule
 \parbox[c]{2mm}{\multirow{8}{*}{\rotatebox[origin=c]{90}{Unseen}}}  & BoolQ&  72.2\err{0.9} & 75.1\err{1.1} &  73.9\err{1.1}& 72.3\err{1.3} & 56.1\err{2.3} & 62.9\err{2.3} \\
 &  NQ&  73.1\err{1.0} & 76.1\err{1.1} &  74.0\err{1.1}& 74.2\err{1.2} & 60.8\err{2.3} & 68.8\err{2.2} \\
 &  MS-Marco&  79.0\err{0.9} & 81.7\err{1.0} &  72.2\err{1.2}& 77.6\err{1.2} & 60.0\err{2.3} & 68.7\err{2.1} \\
 & OBQA& 77.4\err{0.9} & 80.3\err{0.9} &  74.5\err{1.1}& 74.6\err{1.2} & 78.9\err{1.9} & 74.9\err{2.0} \\
 & MMLU-Med& 73.7\err{0.9} & 77.4\err{1.0} &  73.9\err{1.1}& 80.0\err{1.1} & 71.4\err{2.1} & 68.7\err{2.1} \\
 & PubMedQA&  79.5\err{0.9} & 84.0\err{1.0} &  70.2\err{1.3}& 81.9\err{1.2} & 59.3\err{2.4} & 61.1\err{2.4} \\
 &  ARC-Easy&  77.5\err{0.8} & 80.7\err{0.9} &  77.1\err{1.1}& 79.0\err{1.2}& 79.2\err{1.9} & 76.3\err{1.9} \\
& \textbf{AVG} & \textbf{76.1} & \textbf{79.4} &  \textbf{73.7} & \textbf{77.1}& \textbf{66.5} & \textbf{68.8}\\
\bottomrule
\end{tabular}
    \caption{Main result: \method{} improves explanation consistency (on both related questions and rephrased questions) and prediction accuracy.
    Error bars show standard error of the mean.
    }
    \label{tab:main_results}
\end{table*}

\begin{table*}
    \centering
    \fontsize{10}{11.5}\selectfont
    
\begin{tabular}{p{1in} p{2.6in} p{2.4in}}
\toprule
Initial Question & Baseline Explanation & \methodshort{} Explanation \\
\midrule
Are they making a season 4 of the flash? & The flash is a TV show that has been on for 3 seasons. TV shows are typically cancelled after 3 seasons. \textbf{Consistency: 0.00} & The CW has renewed The Flash for a seventh season. \textbf{Consistency: 0.74} \\
\midrule
Does age of ultron come after winter soldier? & Winter soldier was released in 2014. Age of Ultron was released in 2015. \textbf{Consistency: 0.06} & Captain America: The Winter Soldier was released in 2014. Avengers: Age of Ultron was released in 2015. \textbf{Consistency: 0.95} \\
\bottomrule
\end{tabular}
    \caption{Examples of more consistent explanations from \method{}. \method{} improves explanation consistency in different ways.}
    \label{tab:explanation-examples}
\end{table*}

We perform \method{} on the LLaMA-2 13-billion parameter model~\cite{touvron2023llama2}. 
For finetuning, we use 4 datasets:
StrategyQA~\cite{geva-etal-2021-aristotle}, MedMCQA~\cite{pal2022medmcqa}, and two versions of MedQA~\cite{zhang2018medical}:
\mbox{MedQA-Sim} contains related questions on diagnosis and treatment (similar to the original questions), whereas MedQA-Diff contains related questions on medical facts derived from the original questions.

We additionally evaluate consistency on 7 datasets that are not used for finetuning:
BoolQ~\cite{clark2019boolq}, Natural Questions (NQ)~\cite{kwiatkowski2019natural}, MS-Marco~\cite{nguyen2016ms}, OBQA~\cite{OpenBookQA2018}, MMLU-Medical~\cite{hendrycks2020measuring}, PubMedQA~\cite{jin2019pubmedqa} and ARC-Easy~\cite{clark2018think}.
For a cleaner evaluation, these 7 datasets are all converted to have a shared yes-no answer format.
We show each dataset's domain and the skills it tests in \cref{tab:datasets}.
The testing datasets introduce a distribution shift as they cover new domains (science) and new skills (commonsense reasoning and quantitative reasoning) not seen during finetuning.
\cref{tab:dataset-statistics} shows the size of each dataset.

\subsection{Main result: \method{} improves explanation consistency}

\cref{tab:main_results} shows the main results for \method.
\method{} can effectively improve consistency, yielding an average relative improvement of  \textbf{10.0\%}
for tasks seen during finetuning and 
\textbf{4.5\%}
for unseen tasks.
An improvement is seen for every dataset studied here and for both types of followup questions.
The largest gain in consistency after \method{} is for MedQA-Diff;
this suggests that \method{} can also improve the LLM's explanation consistency on 
related questions that are more different 
from the original questions.
These consistency improvements also come with modest accuracy improvements (\textbf{5.2\%} relative for finetuning tasks and \textbf{4.3\%} relative for unseen tasks).
There is no significant correlation between improvement in consistency and the improvement in accuracy (Pearson correlation coefficient $\rho=0.001$).
This suggests that the consistency improvement derived from \method{} differs from the improvement attained by standard supervised finetuning.


\paragraph{\method{} using only a single LLM}
We explore a simplified setting, where \method{} is run using only the LLaMA-2 13-billion parameter, both for synthetic data generation and explanation finetuning.
This setting tests whether \method{} can be used with smaller LLMs and whether those LLMs can improve their own explanation consistency.
We find that when running \method{} on StrategyQA, \method{} yields a 4.4\% relative improvement but decreases accuracy by 5.4\%.
This suggests that \method{} may succeed in improving explanation consistency in today's relatively small models, but can incur some tradeoffs as a result, i.e. decreasing accuracy.

\begin{table}[t]
    \fontsize{10}{11.5}\selectfont
    \centering
    \begin{tabular}{lll}
\toprule
& Baseline & \methodshort\\
\midrule
BoolQ & 0.047 & 0.232 \\
PubMedQA & -0.048 & 0.041 \\
NQ & 0.138 & 0.297 \\
MS-Marco & 0.133 & 0.267 \\
ARC-Easy & 0.152 & 0.129 \\
OBQA & 0.219 & 0.264 \\
MMLU-Medical & 0.050 & 0.069 \\
\textbf{AVG} & \textbf{0.099} & \textbf{0.185}\\
\bottomrule
\end{tabular}
    \caption{Pearson Correlation between explanation consistency and prediction accuracy across examples. Inconsistent explanations suggest wrong predictions.}
    \label{tab:consistency-acc-corr}
\end{table}

\begin{table}[ht]
    \fontsize{10}{11.5}\selectfont
    \centering
    \begin{tabular}{lcc}
\toprule
& Correct & Incorrect\\
\midrule
BoolQ & 5.2\% & 0.0\% \\
NQ & 8.3\% & -5.6\% \\
MS-Marco & 5.9\% & -2.8\% \\
OBQA & 4.9\% & 1.6\% \\
MMLU-Medical & 4.8\% & 6.5\% \\
PubMedQA & 5.5\% & 5.1\% \\
ARC-Easy & 4.8\% & 3.6\% \\
\textbf{AVG} & \textbf{5.8\% }& \textbf{0.8\%} \\
\bottomrule
\end{tabular}
    \caption{Consistency improvement from \methodshort{} on correct vs. incorrect predictions. \method{} improves explanation consistency more on correct predictions.}
    \label{tab:consistency-improvement}
\end{table} 

\subsection{Analysis}

\paragraph{\method{} improves explanation consistency in different ways.}
\cref{tab:explanation-examples} shows examples of explanations before/after \method. The consistency of the explanation in both examples increases after \method, but in different ways. In the first example, \method{} encourages the model to generate more precise explanations that are not overgeneralized/vague. On the other hand, in the second example, \method{} does not change the explanation the model generates for the initial question, but instead changes the model's predictions on related questions to be more consistent with the explanation on the initial question.

\paragraph{Inconsistent explanations suggest incorrect predictions.}
Do LLMs generate more consistent explanations on correct predictions? We study the correlation between explanation consistency and prediction accuracy across different examples of the same dataset.
We find that the baseline model shows a positive correlation of 0.099 (Pearson), and this correlation increases to 0.185 after \method{} (dataset-level breakdown in \cref{tab:consistency-acc-corr}).
This indicates that inconsistent explanations suggest wrong predictions, and we may calibrate LM's predictions based on the consistency of its explanations \cite{chen-etal-2023-relation}.

\paragraph{\method{} improves consistency more on correct predictions.}
We compare the consistency improvement from \method{} on correct versus incorrect predictions.
\method{} improves explanation consistency on correct predictions by 5.7\% relative but only 1.2\% relative on incorrect predictions (see full breakdown in \cref{tab:consistency-improvement}).
This is possibly because examples where the model makes mistakes are likely challenging ones beyond the model's capability, so the model can produce inconsistent hallucinations.

\section{Conclusion}

Forming reliable mental models of LLMs will be critical to understanding and deploying them as their capabilities continue to grow.
\method{} is a step in this direction, helping to align models to not only be more accurate, but also to provide more self-consistent explanations.
A big part of future work will be \emph{i)} scaling up \method{} to larger models, \emph{ii)} exploring \method{} on more complicated tasks (e.g., multimodal understanding), \emph{iii)} studying how LLMs represent consistency in their parameters and if \method{} improves this representation, and \emph{iv)} proposing other methods to improve explanation consistency.



\section{Acknowledgements}
This research is supported in part by Open Philanthropy.
We thank Kathleen McKeown, Zhou Yu, Sarah Wiegreffe, Ruiqi Zhong and Qing Lyu for valuable discussions and feedback on this paper.
\FloatBarrier
{
    \small
    \bibliography{refs}
}

\clearpage
\appendix

\section{Appendix}
\FloatBarrier
\renewcommand{\thefigure}{A\arabic{figure}}
\renewcommand{\thetable}{A\arabic{table}}
\setcounter{table}{0}
\setcounter{figure}{0}

\begin{table}
    \centering
    \fontsize{10}{11.5}\selectfont
    \begin{tabular}{lll}
\toprule
 & Baseline & EC \\
\midrule
BoolQ & -0.059 & 0.039 \\
PubMedQA & 0.044 & 0.007 \\
NQ & -0.002 & 0.028 \\
MS-Marco & 0.084 & 0.006\\
ARC-Easy & 0.030 & 0.012\\
OBQA & -0.066 & -0.034 \\
MMLU-Med & -0.020 & 0.003 \\
\bottomrule
\end{tabular}

    \caption{We find no significant correlation ($p$-value $<0.005$) between explanation consistency and explanation length for either the baseline or EC.
    }
    \label{tab:consistency_length_correlation}
\end{table}

\begin{table}
    \centering
    \fontsize{10}{11.5}\selectfont
    \begin{tabular}{lllll}
\toprule
& & Train & Dev & \multicolumn{1}{l}{Test} \\
\midrule
\multirow{3}{*}{\rotatebox[origin=c]{90}{FT}} & StrategyQA & 1400 & 100 & 790 \\
&MedQA & 2498 & 100 & 600 \\
&MedMCQA & 2500 & 100 & 600 \\
\midrule
\multirow{7}{*}{\rotatebox[origin=c]{90}{Unseen}} 
& BoolQ & -- & -- & 500 \\
&NQ & -- & -- & 500 \\
&MS-Marco & -- & -- & 500 \\
&OBQA & -- & -- & 500 \\
&MMLU-Medical & -- & -- & 500 \\
&PubMedQA & -- & -- & 445 \\
&ARC-Easy & -- & -- & 500 \\
\bottomrule
\end{tabular}

    \caption{Sizes of all datasets used.
    }
    \label{tab:dataset-statistics}
\end{table}

\begin{table*}
    \centering
    \fontsize{10}{11.5}\selectfont
    
\begin{tabular}{llllcl}
\toprule
\multicolumn{2}{c}{Explanation Format} & \multicolumn{3}{c}{Followup Question Generation} & \multicolumn{1}{c}{\multirow{2}{*}{Consistency}} \\
\cmidrule[0.2pt](l{2pt}r{2pt}){1-2} \cmidrule[0.2pt](l{2pt}r{2pt}){3-5}
Brevity & Includes Answer & Model & Decoding & \multicolumn{1}{l}{Prompt} & \multicolumn{1}{c}{} \\
\midrule
Facts & Yes & GPT-4 & List & A & 84.8\err{2.0} \\
Facts & Yes & GPT-4 & List & B & 88.8\err{1.8} \\
Facts & Yes & GPT-4 & Sampling & A & 85.1\err{2.3} \\
Facts & Yes & GPT-4 & Sampling & B & 87.6\err{2.0} \\
Facts & No & GPT-4 & List & A & 85.7\err{2.1} \\
Facts & No & GPT-4 & List & B & 85.9\err{2.3} \\
Facts & No & GPT-4 & Sampling & A & 82.9\err{3.1} \\
Facts & No & GPT-4 & Sampling & B & 84.9\err{2.5} \\
Facts+Reasoning & Yes & GPT-4 & List & A & 85.6\err{2.1} \\
Facts+Reasoning & Yes & GPT-4 & List & B & 85.1\err{2.0} \\
Facts+Reasoning & Yes & GPT-4 & Sampling & A & 87.7\err{2.0} \\
Facts+Reasoning & Yes & GPT-4 & Sampling & B & 84.9\err{2.6} \\
Facts+Reasoning & No & GPT-4 & List & A & 85.5\err{2.1} \\
Facts+Reasoning & Yes & GPT-3 & Sampling & A & 83.2\err{2.3} \\
\bottomrule
\end{tabular}
    \caption{We evaluate the stability of the consistency metric with respect to how the metric is computed. We run several perturbations for the explanation format (whether the explanation includes the reasoning chain besides listing relevant facts, and whether the explanation includes the answer) and how the followup questions are generated (model, decoding, and prompt variations). We find that the consistency metric is quite stable with respect to how the metric is computed.
    }
    \label{tab:consistency_stability}
\end{table*}

\subsection{Prompting details}
\label{sec:prompting_details}
We prompt LLMs to generate the EC training data in two steps 1) generate follow-up questions related to an initial explanation, 2) answer the related questions conditioned on the initial explanation. We include a few demonstration examples for both steps (in-context learning), which are sampled from each dataset. We include below the prompts we use for the StrategyQA dataset, and include the demonstration examples for other datasets at \url{https://github.com/yandachen/explanation-consistency-finetuning}.

\paragraph{Follow-up Questions Generation}
Human: In the questions below, you will be asked to read a starter question and its answer. After that you will be asked to write a follow-up question that can be answered based on the starter QA, and write your answer to the follow-up question based on the starter QA. Your follow-up question should be self-contained even without the starter question.

Assistant: here is my response. okay.

Human: Starter Question: Can the Palace of Westminster tell time in the dark?
Answer to the Starter Question: The Palace of Westminster is a building. Buildings cannot tell time.

Assistant: here is my response. Follow-up Question: Can the White House tell time?
Answer to the Follow-up Question: We know from the starter QA that buildings cannot tell time. Since the White House is also a building, the White House cannot tell time either. So the answer is no.

Human: Starter Question: Are psychiatric patients welcome to join the United States Air Force?
Answer to the Starter Question: The United States Air Force is a branch of the United States military. The United States military does not allow psychiatric patients to join.

Assistant: here is my response. Follow-up Question: Are psychiatric patients welcome to join the United States Army?
Answer to the Follow-up Question: We know from the starter QA that the United States military does not allow psychiatric patients to join. Because United States Army is a branch of the United States military, psychiatric patients are not welcome to join the United States Army either. So the answer is no.

Human: Starter Question: Does The Hague border multiple bodies of water?
Answer to the Starter Question: The Hague is a city in the Netherlands. The Netherlands borders the North Sea.

Assistant: here is my response. Follow-up Question: Is Amsterdam near the North Sea?
Answer to the Follow-up Question: We know from the starter QA that the Netherlands borders the North Sea. Because Amsterdam is a city in the Netherlands, Amsterdam is also near the North Sea. So the answer is yes.

Human: Starter Question: Can an Asian black bear use chopsticks?
Answer to the Starter Question: Asian black bears are omnivores. Omnivores can use chopsticks.

Assistant: here is my response. Follow-up Question: Can pigs use chopsticks?
Answer to the Follow-up Question: We know from the starter QA that omnivores can use chopsticks. Because pigs are omnivores, pigs can use chopsticks as well. So the answer is yes.

Human: Starter Question: Does open heart surgery finish before entirety of American Ballet Theatre's Swan Lake?
Answer to the Starter Question: The American Ballet theatre's Swan Lake has a run time of 45 minutes. The National Heart, Lung, and Blood Institute states that a coronary artery bypass takes 3 to 6 hours.

Assistant: here is my response. Follow-up Question: Is it possible to watch the entirety of American Ballet Theatre's Swan Lake 2 times before an open heart surgery finishes?
Answer to the Follow-up Question: We know from the starter QA that the run time of Swan Lake is 45 minutes and an open heart surgery takes 3 to 6 hours. Therefore it is possible to watch Swan Lake twice before the surgery finishes (90 minutes). So the answer is yes.

Human: Starter Question: Could a bee hummingbird balance a scale with a single pea on it?
Answer to the Starter Question: The bee hummingbird is the smallest bird in the world. The bee hummingbird weighs about 2 grams. A pea weighs about 1 gram.

Assistant: here is my response. Follow-up Question: Could a pea balance a scale with a dollar bill on it?
Answer to the Follow-up Question: We know from the starter QA that a pea weighs about 1 gram. Because a dollar bill weighs about 1 gram as well, a pea can balance a scale with a dollar bill on it. So the answer is yes.

Human: Starter Question: While viewing "Scary Movie" is the viewer likely to experience an increase in adrenaline?
Answer to the Starter Question: The movie "Scary Movie" is a comedy. Comedies are not scary.

Assistant: here is my response. Follow-up Question: Is "Superbad" scary?
Answer to the Follow-up Question: We know from the starter QA that comedies are not scary. "Superbad" is also a comedy, so "Superbad" is not scary either. So the answer is no.

Human: Starter Question: Is Mixed martial arts totally original from Roman Colosseum games?
Answer to the Starter Question: Mixed Martial arts in the UFC takes place in an enclosed structure called The Octagon. The Roman Colosseum games were fought in enclosed arenas where combatants would fight until the last man was standing. Mixed martial arts contests are stopped when one of the combatants is incapacitated. The Roman Colosseum was performed in front of crowds that numbered in the tens of thousands. Over 56,000 people attended UFC 193.

Write 10 follow-up yes or no questions. Start each question by "1. ", "2. ", etc. Skip writing answers to your follow-up questions. Remember, your task is to write follow-up questions that can be answered based on the starter QA. Your follow-up questions should be self-contained even without the starter question.

\paragraph{Follow-up Questions Answering}
Human: In the questions below, you will be asked to read a starter question and its answer. After that you will be asked to read a follow-up question and judge whether the starter QA directly helps choosing a single answer for the follow-up question. If not, end your answer with "So the answer is unknown.". If yes, use the starter QA to answer the follow-up question, explain your reasoning as clearly and as detailed as possible using all relevant information in the starter QA, end your answer with "So the answer is yes/no.", and do NOT explicitly mention "the starter QA" or "According to the starter QA" in your answer. Stick to the starter QA when you answer the follow-up question, even if the reasoning or claims in the starter QA are wrong.

Assistant: here is my response. okay.

Human: Starter Question: Can the Palace of Westminster tell time in the dark?
Answer to the Starter Question: The Palace of Westminster is a building. Buildings cannot tell time.
Follow-up Question: Can the White House tell time?
Judge whether the starter QA directly helps choosing a single answer for the follow-up question. If not, end your answer with "So the answer is unknown.". If yes, use the starter QA to answer the follow-up question, explain your reasoning as clearly and as detailed as possible using all relevant information in the starter QA, end your answer with "So the answer is yes/no.", and do NOT explicitly mention "the starter QA" or "According to the starter QA" in your answer. Stick to the starter QA when you answer the follow-up question, even if the reasoning or claims in the starter QA are wrong.
Answer to the Follow-up Question:

Assistant: here is my response. The White House is a building. Buildings cannot tell time. So the answer is no.

Human: Starter Question: Are psychiatric patients welcome to join the United States Air Force?
Answer to the Starter Question: The United States Air Force is a branch of the United States military. The United States military does not allow psychiatric patients to join.
Follow-up Question: Are psychiatric patients welcome to join the United States Army?
Judge whether the starter QA directly helps choosing a single answer for the follow-up question. If not, end your answer with "So the answer is unknown.". If yes, use the starter QA to answer the follow-up question, explain your reasoning as clearly and as detailed as possible using all relevant information in the starter QA, end your answer with "So the answer is yes/no.", and do NOT explicitly mention "the starter QA" or "According to the starter QA" in your answer. Stick to the starter QA when you answer the follow-up question, even if the reasoning or claims in the starter QA are wrong.
Answer to the Follow-up Question:

Assistant: here is my response. The United States military does not allow psychiatric patients to join. The United States Army is a branch of the United States military. So the answer is no.

Human: Starter Question: Does The Hague border multiple bodies of water?
Answer to the Starter Question: The Hague is a city in the Netherlands. The Netherlands borders the North Sea.
Follow-up Question: Is Amsterdam near the North Sea?
Judge whether the starter QA directly helps choosing a single answer for the follow-up question. If not, end your answer with "So the answer is unknown.". If yes, use the starter QA to answer the follow-up question, explain your reasoning as clearly and as detailed as possible using all relevant information in the starter QA, end your answer with "So the answer is yes/no.", and do NOT explicitly mention "the starter QA" or "According to the starter QA" in your answer. Stick to the starter QA when you answer the follow-up question, even if the reasoning or claims in the starter QA are wrong.
Answer to the Follow-up Question:

Assistant: here is my response. Amsterdam is a city in the Netherlands. Netherlands borders the North Sea. So the answer is yes.

Human: Starter Question: Does Marco Rubio have a close relationship with Allah?
Answer to the Starter Question: Marco Rubio is a Christian. Christians do not have a close relationship with Allah.
Follow-up Question: Is Barack Obama a Christian?
Judge whether the starter QA directly helps choosing a single answer for the follow-up question. If not, end your answer with "So the answer is unknown.". If yes, use the starter QA to answer the follow-up question, explain your reasoning as clearly and as detailed as possible using all relevant information in the starter QA, end your answer with "So the answer is yes/no.", and do NOT explicitly mention "the starter QA" or "According to the starter QA" in your answer. Stick to the starter QA when you answer the follow-up question, even if the reasoning or claims in the starter QA are wrong.
Answer to the Follow-up Question:

Assistant: here is my response. The starter QA does not include information that directly answer this follow-up question. So the answer is unknown.

Human: Starter Question: Can an Asian black bear use chopsticks?
Answer to the Starter Question: Asian black bears are omnivores. Omnivores can use chopsticks.
Follow-up Question: Can pigs use chopsticks?
Judge whether the starter QA directly helps choosing a single answer for the follow-up question. If not, end your answer with "So the answer is unknown.". If yes, use the starter QA to answer the follow-up question, explain your reasoning as clearly and as detailed as possible using all relevant information in the starter QA, end your answer with "So the answer is yes/no.", and do NOT explicitly mention "the starter QA" or "According to the starter QA" in your answer. Stick to the starter QA when you answer the follow-up question, even if the reasoning or claims in the starter QA are wrong.
Answer to the Follow-up Question:

Assistant: here is my response. Pigs are omnivores. Omnivores can use chopsticks. So the answer is yes.

Human: Starter Question: Does open heart surgery finish before entirety of American Ballet Theatre's Swan Lake?
Answer to the Starter Question: The American Ballet theatre's Swan Lake has a run time of 45 minutes. The National Heart, Lung, and Blood Institute states that a coronary artery bypass takes 3 to 6 hours.
Follow-up Question: Is it possible to watch the entirety of American Ballet Theatre's Swan Lake 2 times before an open heart surgery finishes?
Judge whether the starter QA directly helps choosing a single answer for the follow-up question. If not, end your answer with "So the answer is unknown.". If yes, use the starter QA to answer the follow-up question, explain your reasoning as clearly and as detailed as possible using all relevant information in the starter QA, end your answer with "So the answer is yes/no.", and do NOT explicitly mention "the starter QA" or "According to the starter QA" in your answer. Stick to the starter QA when you answer the follow-up question, even if the reasoning or claims in the starter QA are wrong.
Answer to the Follow-up Question:

Assistant: here is my response. The run time of Swan Lake is 45 minutes. An open heart surgery takes 3 to 6 hours. 45 minutes * 2 = 90 minutes < 3 hours. So the answer is yes.

Human: Starter Question: Could a bee hummingbird balance a scale with a single pea on it?
Answer to the Starter Question: The bee hummingbird is the smallest bird in the world. The bee hummingbird weighs about 2 grams. A pea weighs about 1 gram.
Follow-up Question: Could a pea balance a scale with a dollar bill on it?
Judge whether the starter QA directly helps choosing a single answer for the follow-up question. If not, end your answer with "So the answer is unknown.". If yes, use the starter QA to answer the follow-up question, explain your reasoning as clearly and as detailed as possible using all relevant information in the starter QA, end your answer with "So the answer is yes/no.", and do NOT explicitly mention "the starter QA" or "According to the starter QA" in your answer. Stick to the starter QA when you answer the follow-up question, even if the reasoning or claims in the starter QA are wrong.
Answer to the Follow-up Question:

Assistant: here is my response. A pea weighs about 1 gram. A dollar bill weighs about 1 gram. So the answer is yes.

Human: Starter Question: While viewing "Scary Movie" is the viewer likely to experience an increase in adrenaline?
Answer to the Starter Question: The movie "Scary Movie" is a comedy. Comedies are not scary.
Follow-up Question: Is "Superbad" scary?
Judge whether the starter QA directly helps choosing a single answer for the follow-up question. If not, end your answer with "So the answer is unknown.". If yes, use the starter QA to answer the follow-up question, explain your reasoning as clearly and as detailed as possible using all relevant information in the starter QA, end your answer with "So the answer is yes/no.", and do NOT explicitly mention "the starter QA" or "According to the starter QA" in your answer. Stick to the starter QA when you answer the follow-up question, even if the reasoning or claims in the starter QA are wrong.
Answer to the Follow-up Question:

Assistant: here is my response. The movie "Superbad" is a comedy. Comedies are not scary. So the answer is no.

Human: Starter Question: Do gorillas fight with panda bears?
Answer to the Starter Question: Gorillas are herbivores. Pandas are herbivores. Herbivores do not fight with other herbivores.
Follow-up Question: Do herbivores fight with carnivores?
Judge whether the starter QA directly helps choosing a single answer for the follow-up question. If not, end your answer with "So the answer is unknown.". If yes, use the starter QA to answer the follow-up question, explain your reasoning as clearly and as detailed as possible using all relevant information in the starter QA, end your answer with "So the answer is yes/no.", and do NOT explicitly mention "the starter QA" or "According to the starter QA" in your answer. Stick to the starter QA when you answer the follow-up question, even if the reasoning or claims in the starter QA are wrong.
Answer to the Follow-up Question:

Assistant: here is my response. The starter QA does not include information that directly answer this follow-up question. So the answer is unknown.

Human: Starter Question: Can citrus grow in Ulaanbaatar?
Answer to the Starter Question: Citrus trees can grow in Ulaanbaatar. Thus, citrus can grow in Ulaanbaatar.
Follow-up Question: Can palm trees grow in Ulaanbaatar?
Judge whether the starter QA directly helps choosing a single answer for the follow-up question. If not, end your answer with "So the answer is unknown.". If yes, use the starter QA to answer the follow-up question, explain your reasoning as clearly and as detailed as possible using all relevant information in the starter QA, end your answer with "So the answer is yes/no.", and do NOT explicitly mention "the starter QA" or "According to the starter QA" in your answer. Stick to the starter QA when you answer the follow-up question, even if the reasoning or claims in the starter QA are wrong.
Answer to the Follow-up Question:

Assistant: here is my response. The starter QA does not include information that directly answer this follow-up question. So the answer is unknown.

Human: Starter Question: Is Mixed martial arts totally original from Roman Colosseum games?
Answer to the Starter Question: Mixed Martial arts in the UFC takes place in an enclosed structure called The Octagon. The Roman Colosseum games were fought in enclosed arenas where combatants would fight until the last man was standing. Mixed martial arts contests are stopped when one of the combatants is incapacitated. The Roman Colosseum was performed in front of crowds that numbered in the tens of thousands. Over 56,000 people attended UFC 193.
Follow-up Question: Is the UFC Octagon considerably smaller than the Roman Colosseum?
Judge whether the starter QA directly helps choosing a single answer for the follow-up question. If not, end your answer with "So the answer is unknown.". If yes, use the starter QA to answer the follow-up question, explain your reasoning as clearly and as detailed as possible using all relevant information in the starter QA, end your answer with "So the answer is yes/no.", and do NOT explicitly mention "the starter QA" or "According to the starter QA" in your answer. Stick to the starter QA when you answer the follow-up question, even if the reasoning or claims in the starter QA are wrong.
Answer to the Follow-up Question:

\end{document}